\documentclass{article}

\usepackage{PRIMEarxiv}
\usepackage[utf8]{inputenc} % allow utf-8 input
\usepackage[T1]{fontenc}    % use 8-bit T1 fonts
\usepackage{hyperref}       % hyperlinks
\usepackage{url}            % simple URL typesetting
\usepackage{booktabs}       % professional-quality tables
\usepackage{amsfonts}       % blackboard math symbols
\usepackage{nicefrac}       % compact symbols for 1/2, etc.
\usepackage{microtype}      % microtypography
\usepackage{lipsum}
\usepackage{fancyhdr}       % header
\usepackage{graphicx}       % graphics
\graphicspath{{media/}}     % organize your images and other figures under media/ folder
\usepackage{float}
\usepackage{subfig}
\usepackage{multirow}
\usepackage[toc,page]{appendix}

\title{Deep Self-supervised Learning with Visualisation for Automatic Gesture Recognition
}

\author{
  Fabien ALLEMAND\\
  Department ARTEMIS\\
  Télécom SudParis\\
  \texttt{fabien.allemand@telecom-sudparis.eu}\\
  \And
  Alessio MAZZELLA\\
  Department ARTEMIS\\
  Télécom SudParis\\
  \texttt{alessio.mazzella@telecom-sudparis.eu}\\
  \And
  Jun VILLETTE\\
  Department ARTEMIS\\
  Télécom SudParis\\
  \texttt{jeanne.villette@telecom-sudparis.eu}\\
  \And
  Decky ASPANDI LATIF\\
  Department ARTEMIS\\
  Télécom SudParis\\
  \texttt{decky.aspandi\_latif@telecom-sudparis.eu}\\
  \And
  Titus ZAHARIA\\
  Department ARTEMIS\\
  Télécom SudParis\\
  \texttt{titus.zaharia@telecom-sudparis.eu}\\
}

\begin{document}
\maketitle

\begin{abstract}
Gesture is an important mean of non-verbal communication, with visual modality allows human to convey information during interaction, facilitating peoples and human-machine interactions. However, it is considered difficult to automatically recognise gestures. In this work, we explore three different means to recognise hand signs using deep learning: supervised learning based methods, self-supervised methods and visualisation based techniques applied to 3D moving skeleton data. Self-supervised learning used to train fully connected, CNN and LSTM method. Then, reconstruction method is applied to unlabelled data in simulated settings using CNN as a backbone where we use the learnt features to perform the prediction in the remaining labelled data. Lastly, Grad-CAM is applied to discover the focus of the models. Our experiments results show that supervised learning method is capable to recognise gesture accurately, with self-supervised learning increasing the accuracy in simulated settings. Finally, Grad-CAM visualisation shows that indeed the models focus on relevant skeleton joints on the associated gesture.
\end{abstract}

% keywords

\section{Introduction}

\begin{figure}[t]
    \centering
    \subfloat[\centering Real time hand tracking for Virtual Reality (VR)]{{\includegraphics[width=8cm]{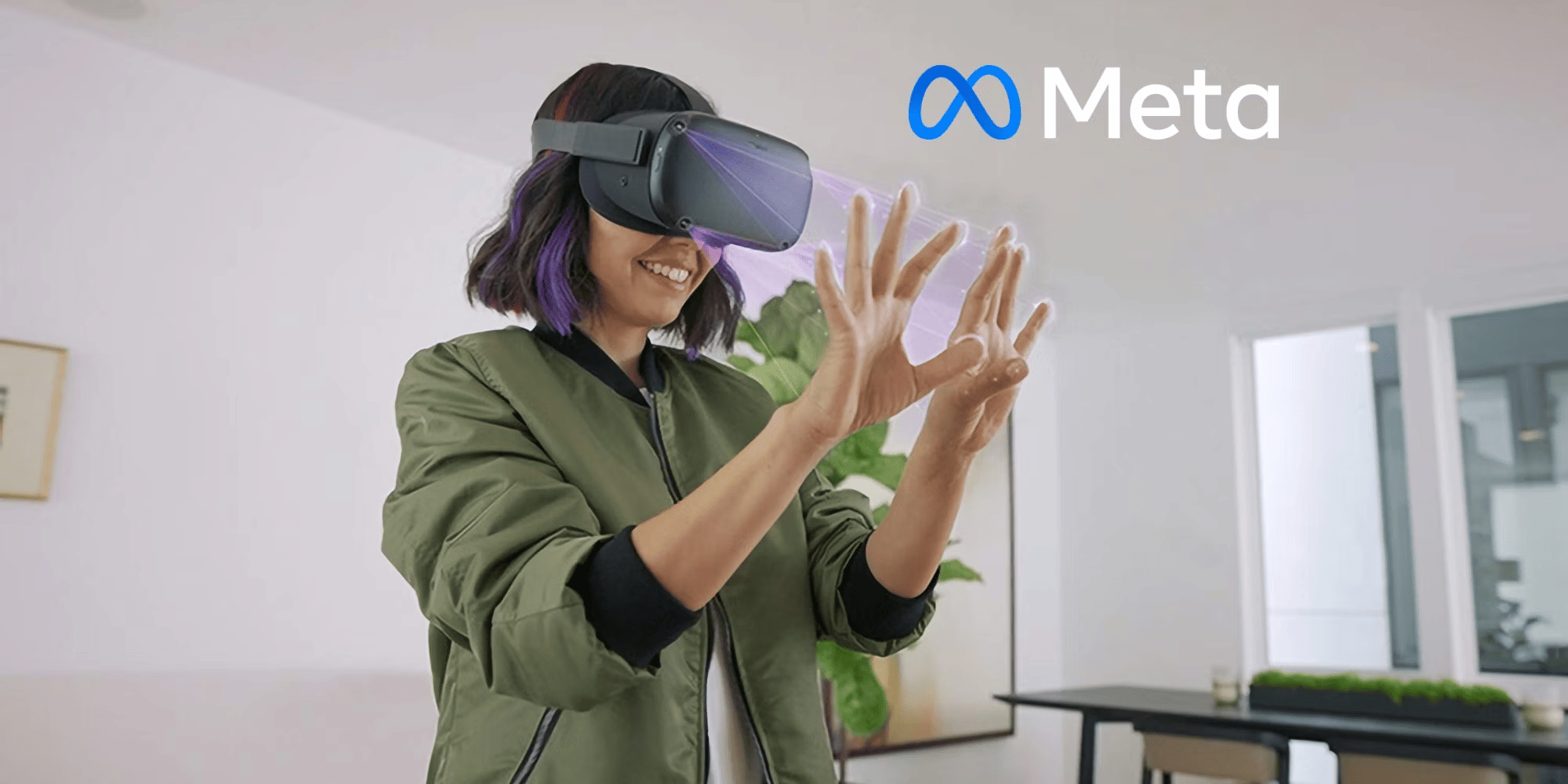} \label{vr}}}
    \qquad
    \subfloat[\centering Sign language recognition for human interactions]{{\includegraphics[width=8cm]{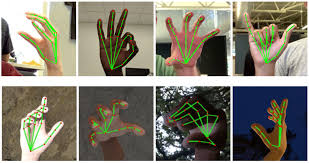} \label{sign_language}}}
    \qquad
    \subfloat[\centering Hand gesture recognition for human to machine communication]{{\includegraphics[width=8cm]{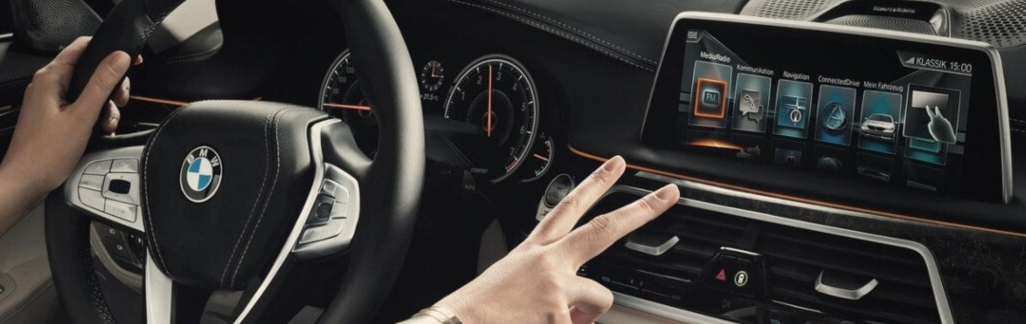} \label{automotive}}}
    \caption{Several applications of gesture recognition.}
    \label{illustrations}
\end{figure}

Gesture is an important means of non-verbal communication facilitating human interactions. Certain conditions of disability or other constraints force some people to communicate using sign languages. However, sign language can also be used for human to machine interactions. In this context, most smart assistants use voice as input, but sign language recognition, as shown in Figure \ref{sign_language}, should be used to enhance accessibility.
 
With the recent development of wearable, virtual reality and spatial computing, hand tracking and gesture recognition have become key method in human-machine communications. Meta offers virtual reality headsets capable of real-time hand tracking (Figure \ref{vr}), and the latest generations of Apple products (Apple Watch and Apple Vision Pro) are attempting to revolutionise the way we interact with our electronic devices using hand gestures.
 
Other applications are to be expected, particularly in the automotive industry, where vehicles are tending to become increasingly intelligent and autonomous thanks to powerful on-board computers connected to numerous sensors. It's not impossible to imagine a vehicle sound or climate control system using gesture recognition as shown in Figure \ref{automotive}. Perhaps one day, there will be possibility of the utilisation of such system similar to the computers in futuristic films, controlled solely by voice and gestures.

In machine learning context, there are several types of automatic learning: Supervised Learning (SL), Unsupervised Learning (UL), reinforcement learning and Self-Supervised Learning (SSL). Although the latter method has proved its worth in other computer vision applications, the work is based on the observation that the use of self-supervised learning methods in the field of automatic gesture recognition is still limited. The aim of this work is therefore to apply a self-supervised learning method using deep neural networks to gesture recognition on 3D skeleton data and to compare the results obtained with a conventional supervised learning method. 

\section{Methodology}

\subsection{Deep Learning for Gesture Recognition}

\begin{figure}[]
    \centering
    \subfloat[\centering Fully Connected (FC) model.]{{\includegraphics[width=10cm]{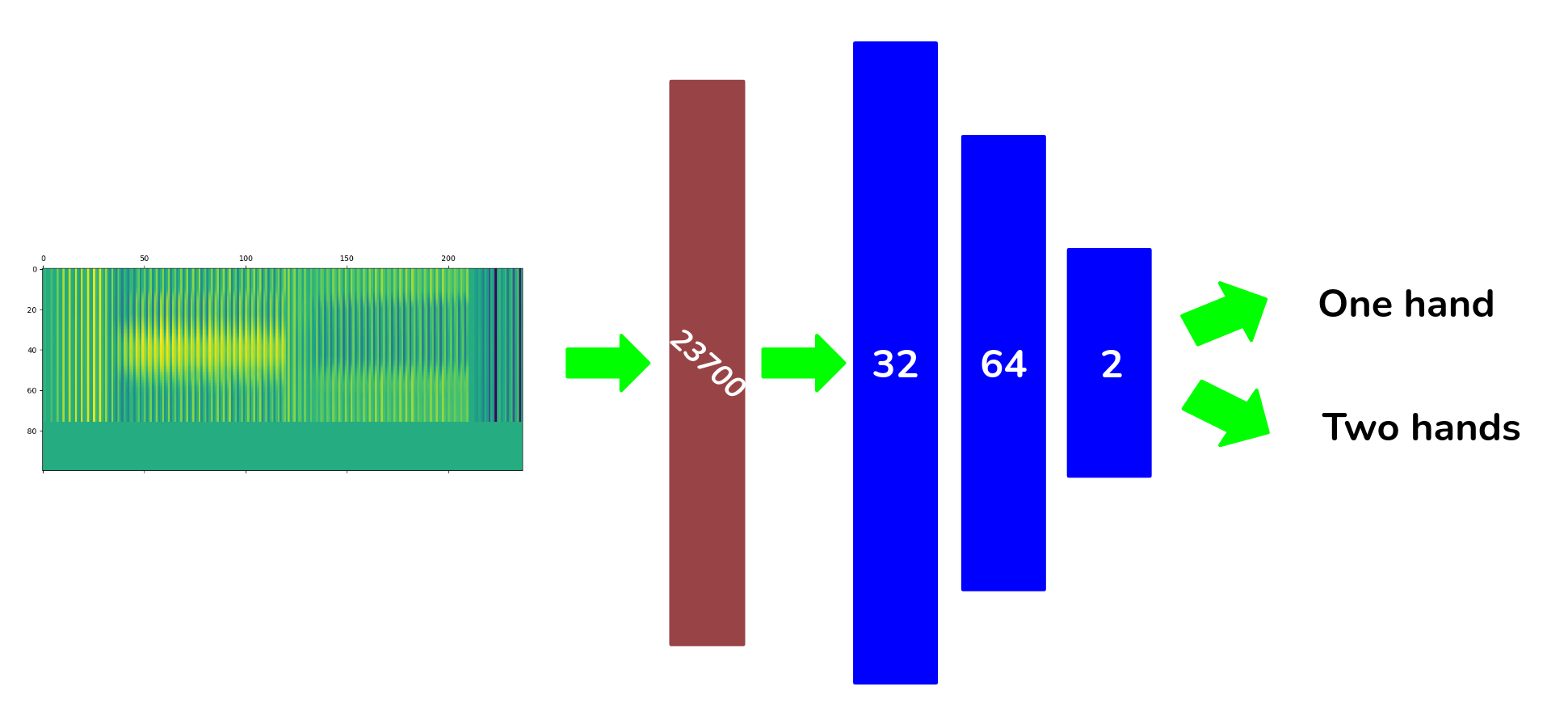} \label{model_FC}}}
    \qquad
    \subfloat[\centering Convolutional Neural Network (CNN) model.]{{\includegraphics[width=10cm]{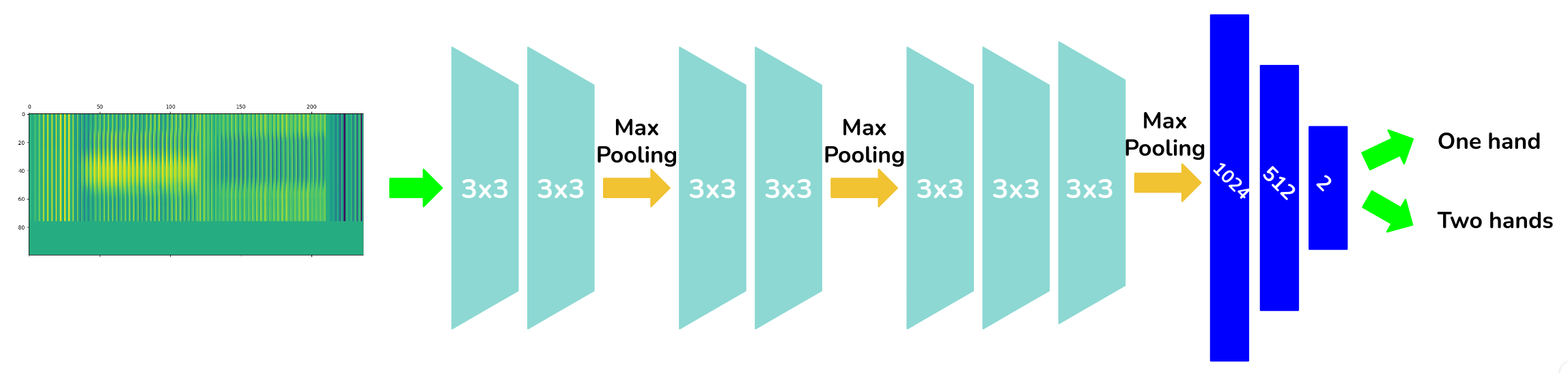} \label{model_CNN}}}
    \qquad
    \subfloat[\centering Long Short-Term Memory (LSTM) model.]{{\includegraphics[width=10cm]{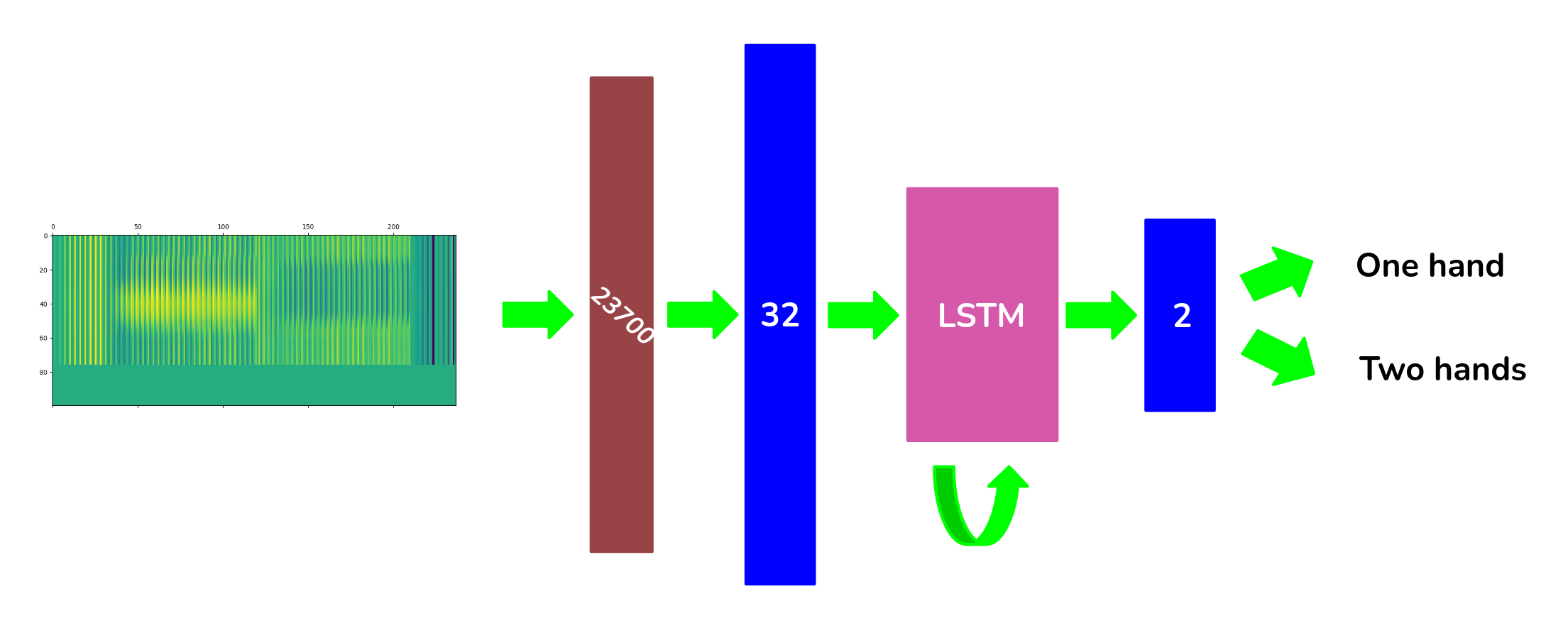} \label{model_LSTM}}}
    \caption{Deep learning models used for sign language recognition.}
    \label{results}
\end{figure}

The goal of our work is to compare the performances of several neural networks architectures for sign language recognition in order to find the type of model that best fit the data containing hands gestures. For this task, we choose three well known neural network architectures and trained then using a supervised approach. Here is a brief description of the three types of neural networks we used: % Add citation
\begin{itemize}
    \item Fully Connected (FC) model: Figure \ref{model_FC} shows the architecture of the FC neural network in blue. There are three layers composed of artificial neurons. Each neuron from a layer is connected to every neuron in the following layer. This model is the most simple one.
    \item Convolutional Neural Network (CNN) model: CNN architecture \cite{cnn_1} is known to perform well in processing two dimensional data such as images as it does not process each value of the input independently. Instead it looks at patches of values allowing it to learn information based on position. The convolution operation used for that learns how to reduce the size of the input data while preserving the most relevant information, in a way, it encodes the input data.  The light blue layers in Figure \ref{model_CNN} depict the convolution layers and the head of the neural network (dark blue layers) are fully connected layers that perform the prediction on the encoded data.
    \item Long Short-Term Memory (LSTM) model: LSTM \cite{lstm_1} are part of a larger family of models called recurrent neural networks. These models are known to work best for input data that corresponds to sequences. The architecture of an LSTM neural network is visualised in Figure \ref{model_LSTM}.
\end{itemize}

\begin{figure}[]
    \centering
    \subfloat[\centering Training step.]{{\includegraphics[width=8cm]{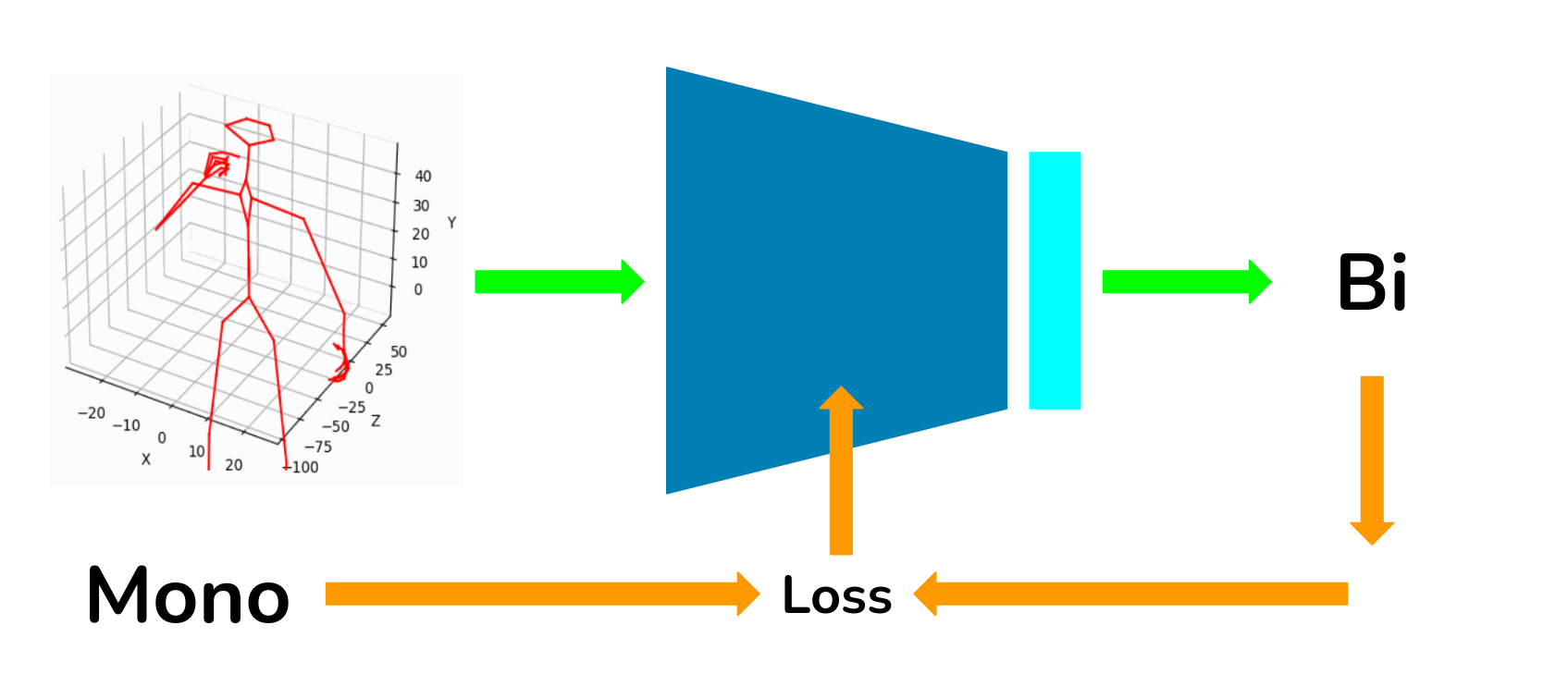} \label{SL_learning}}}
    % \qquad
    \subfloat[\centering Prediction step.]{{\includegraphics[width=8cm]{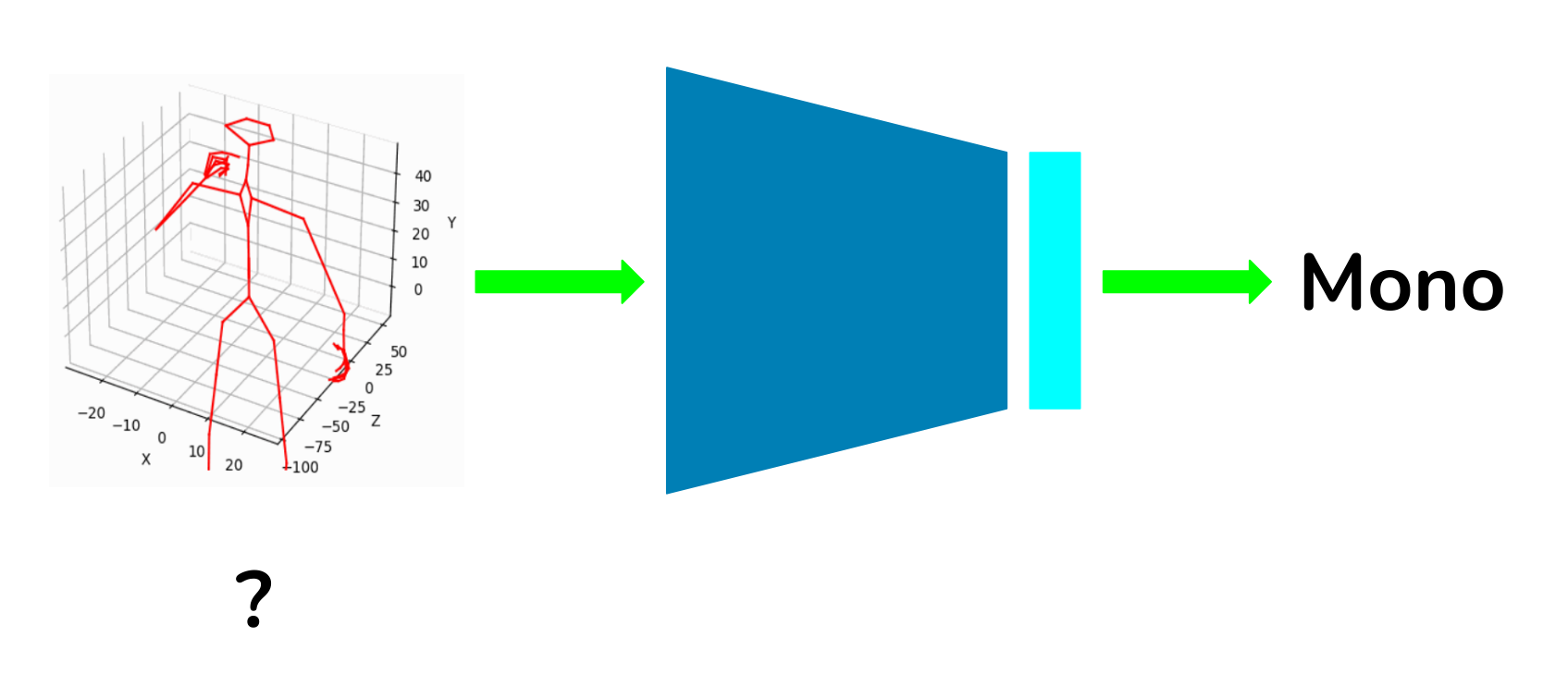} \label{SL_prediction}}}
    \caption{Supervised learning process for class prediction.}
    \label{SL}
\end{figure}

Supervised learning is the most widespread machine learning technique. It consists of the utilisation of the data with labels in a feedback loop that allow the neural network to learn how to best predict the correct label given an input sampled from the data (see Figure \ref{SL_learning}). 

The learning process may vary a little between two neural network architectures but it always consists in adjusting the weights of the neural network in order to minimize the loss: a measure telling how far the prediction is from being correct. After the training, the model can be used to make accurate predictions as shown in Figure \ref{SL_prediction}.

\subsection{Self-supervised Learning}
\label{SSL_section}

An other aspect of this work is to compare the performances of the aforementioned neural networks when trained with regular supervised techniques and with self-supervised techniques, a learning approach that is relevant when dealing with a lot of unlabelled data (as it could be the case with hand gesture) but has never been used for gesture recognition.

\begin{figure}[]
    \centering
    \subfloat[\centering Training step.]{{\includegraphics[width=8cm]{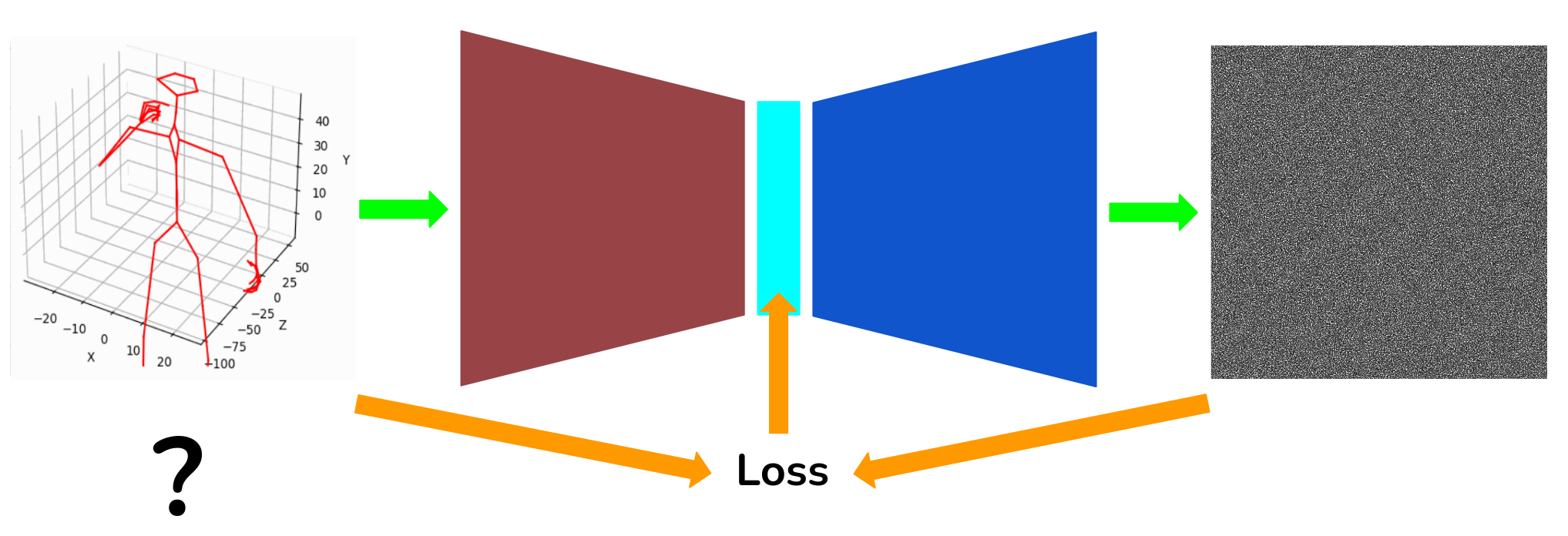} \label{UL_learning}}}
    % \qquad
    \subfloat[\centering Reconstruction step.]{{\includegraphics[width=8cm]{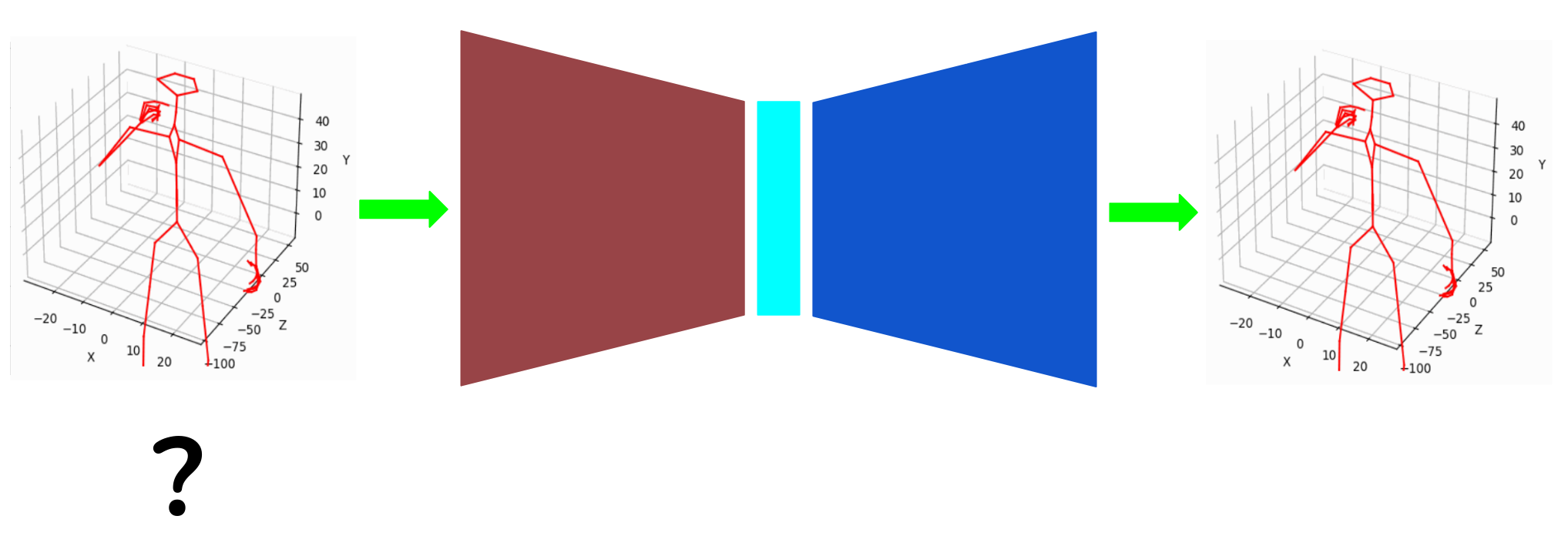} \label{UL_reconstruction}}}
    \caption{Unsupervised learning process for input reconstruction.}
    \label{UL}
\end{figure}

\begin{figure}[]
    \centering
    \includegraphics[width=10cm]{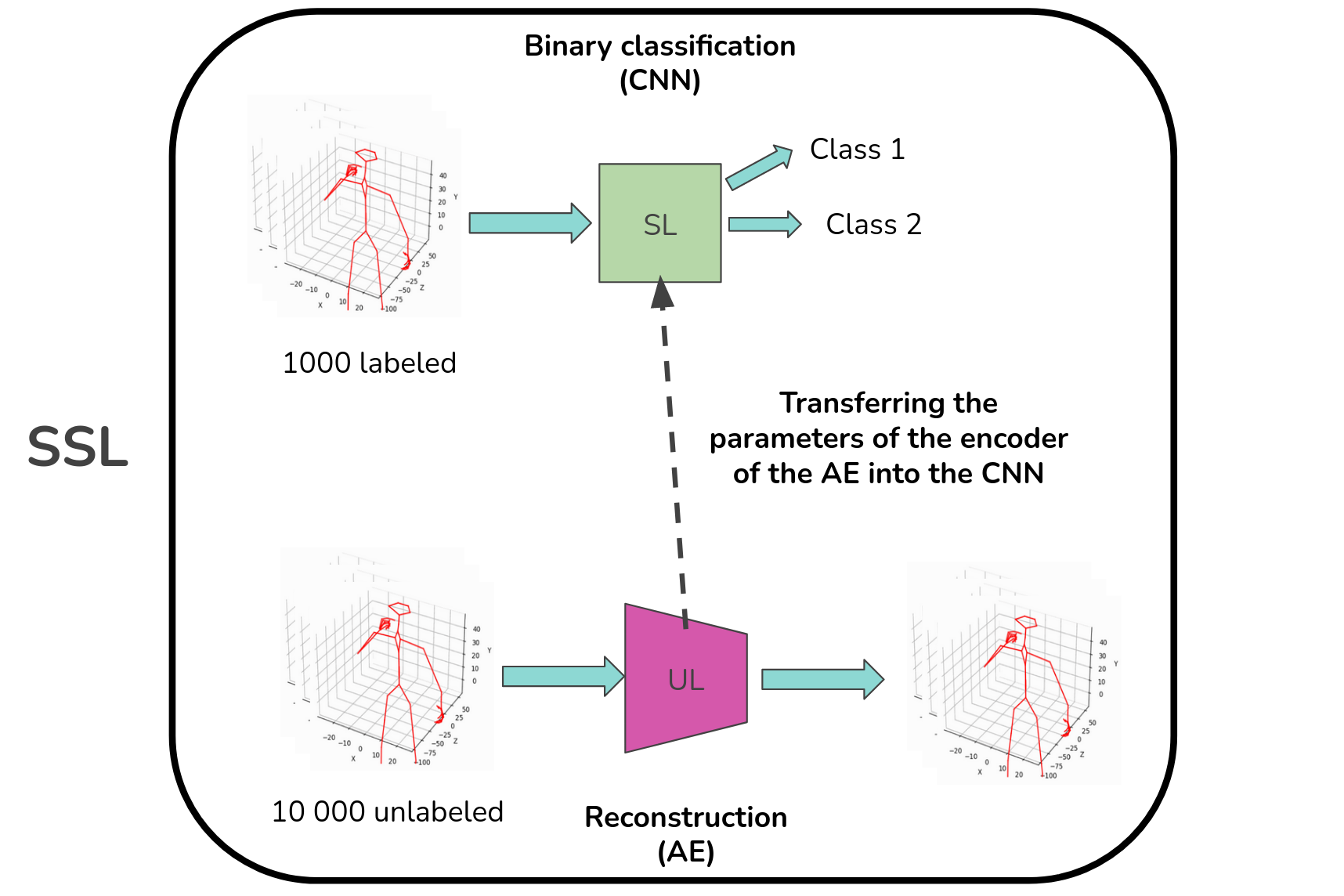}
    \caption{Overview of self-supervised learning for class prediction.}
    \label{SSL_overview}
\end{figure}

Self-supervised learning consists in pre-training the supervised model (that requires expensive labelled data) with easily accessible unlabelled data\cite{schneider2022detecting}. This can be done by training a model following an unsupervised method (such as input reconstruction as explained in Figure \ref{UL}) requiring unlabelled data then taking a relevant part of this model and pursue its training with labelled data. The overall process is summarized in Figure \ref{SSL_overview}. This process consists of training the model to produce latent features~\cite{aspandi2020latent,aspandi2022audio} in unsupervised way (i.e. in this case is reconstruction as pretext task). Then, the trained models used further on downstream task - supervised gesture recognition.

\begin{figure}[]
    \centering
    \includegraphics[width=10cm]{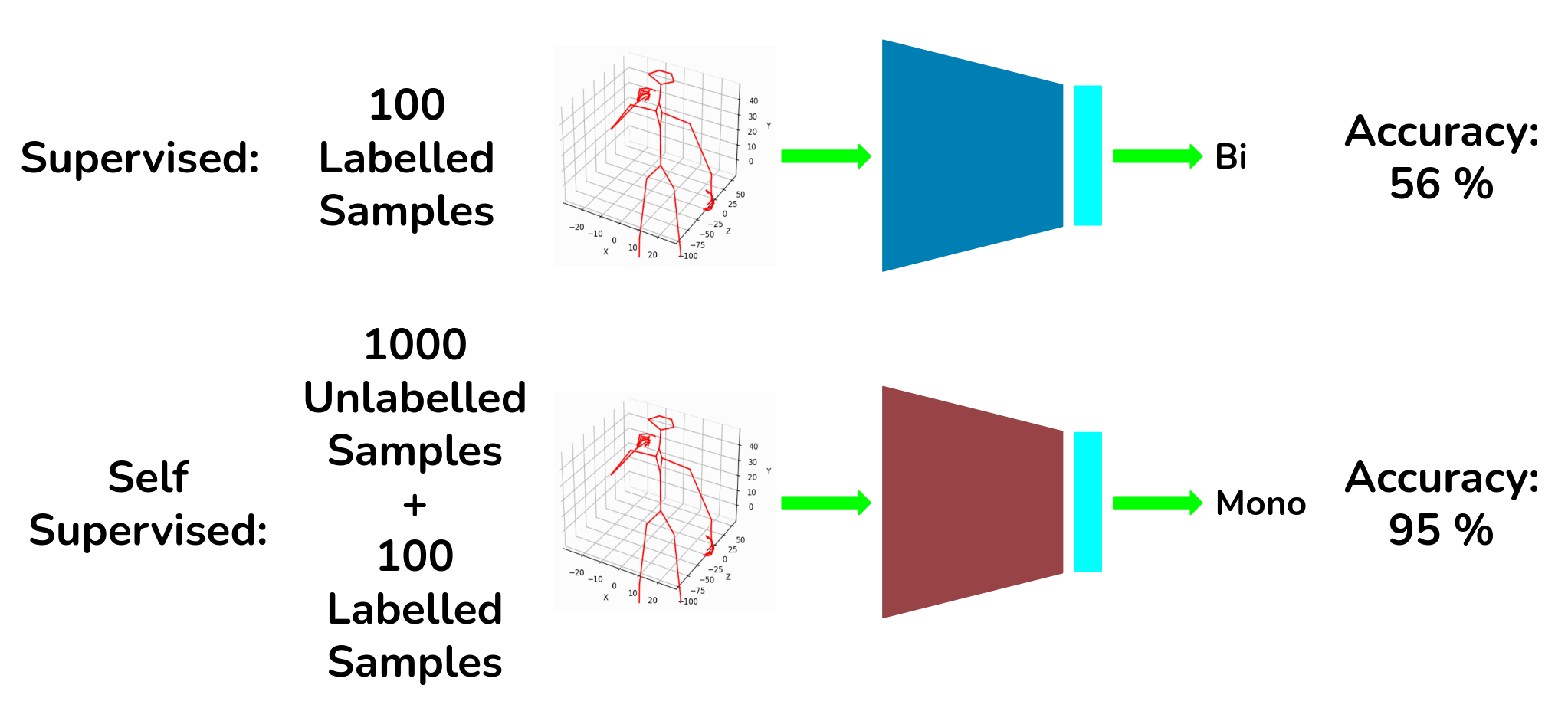}
    \caption{Supervised versus self-supervised learning, expected results (the values are for illustration purpose).}
    \label{expected_results}
\end{figure}

By doing so, we expect the unsupervised learning to help the model understand the data it is working with and the supervised learning to be more efficient. That is to say, with a fixed amount of labelled data, we expect the self-supervised model to be more accurate than the standard supervised learning model thanks to the unsupervised pre-training. Expected results are presented in Figure \ref{expected_results}.

\begin{figure}[]
    \centering
    \subfloat[\centering Contrastive learning.]{{\includegraphics[width=8cm]{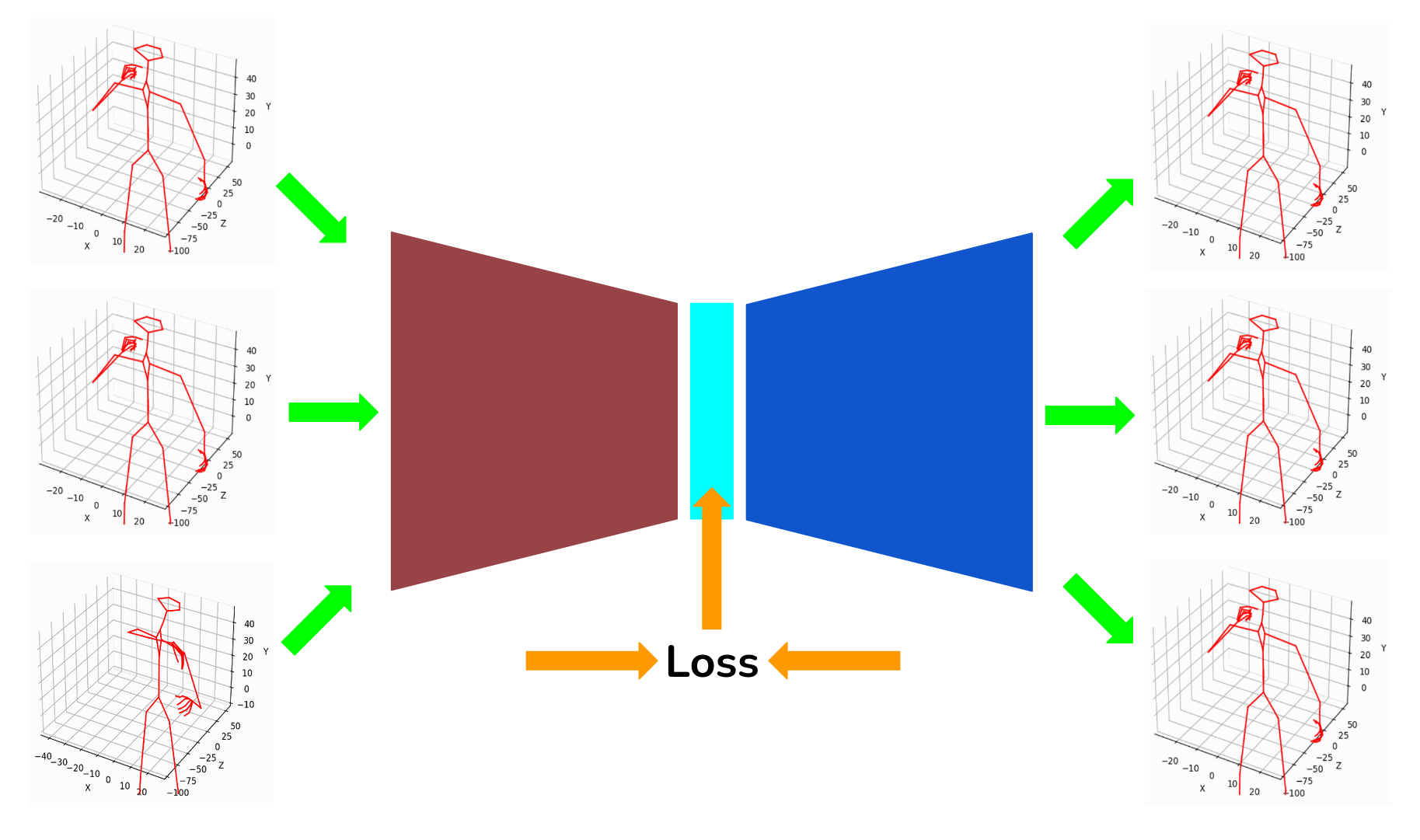}} \label{contrastive_learning}}
    % \qquad
    \subfloat[\centering Contrastive learning expected results (visualisation in the latent space).]{{\includegraphics[width=8cm]{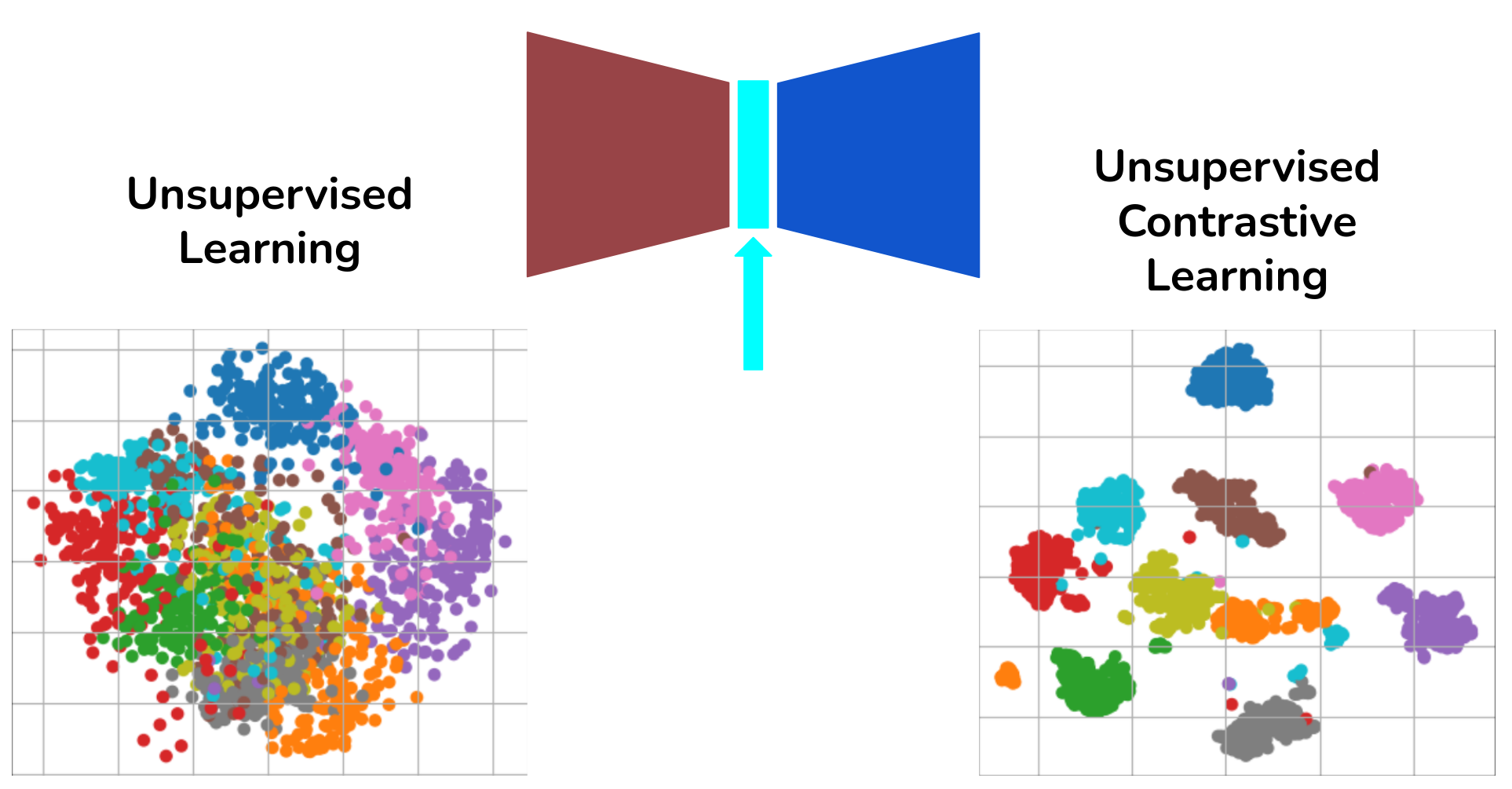} \label{contrastive_expected_results}}}
    \caption{Contrastive learning overview.}
    \label{contrastive}
\end{figure}

Self-supervised learning can also benefit from contrastive learning as demonstrated in \cite{chen2020simple}. This technique allows the model to learn a representation of an input sample in a latent space based on its similarity and dissimilarity with other samples as presented in Figure \ref{contrastive_learning}. Using contrastive learning should help the model to learn a representation of the data in the latent space where similar data are closer together as depicted in Figure \ref{contrastive_expected_results}. We expect that this coherence in the model will help it perform better during the downstream task.

\subsection{Grad-CAM}

visualisation is an essential part of training deep learning models, especially when working on new types of data. It deepens our understanding of the way the models learn and can help explain their observed behaviors and decisions. That is why we applied the Gradient-weighted Class Activation Mapping (Grad-CAM) algorithm \cite{selvaraju2017grad} to our CNN model.

\begin{figure}[]
    \centering
    \includegraphics[width=15cm]{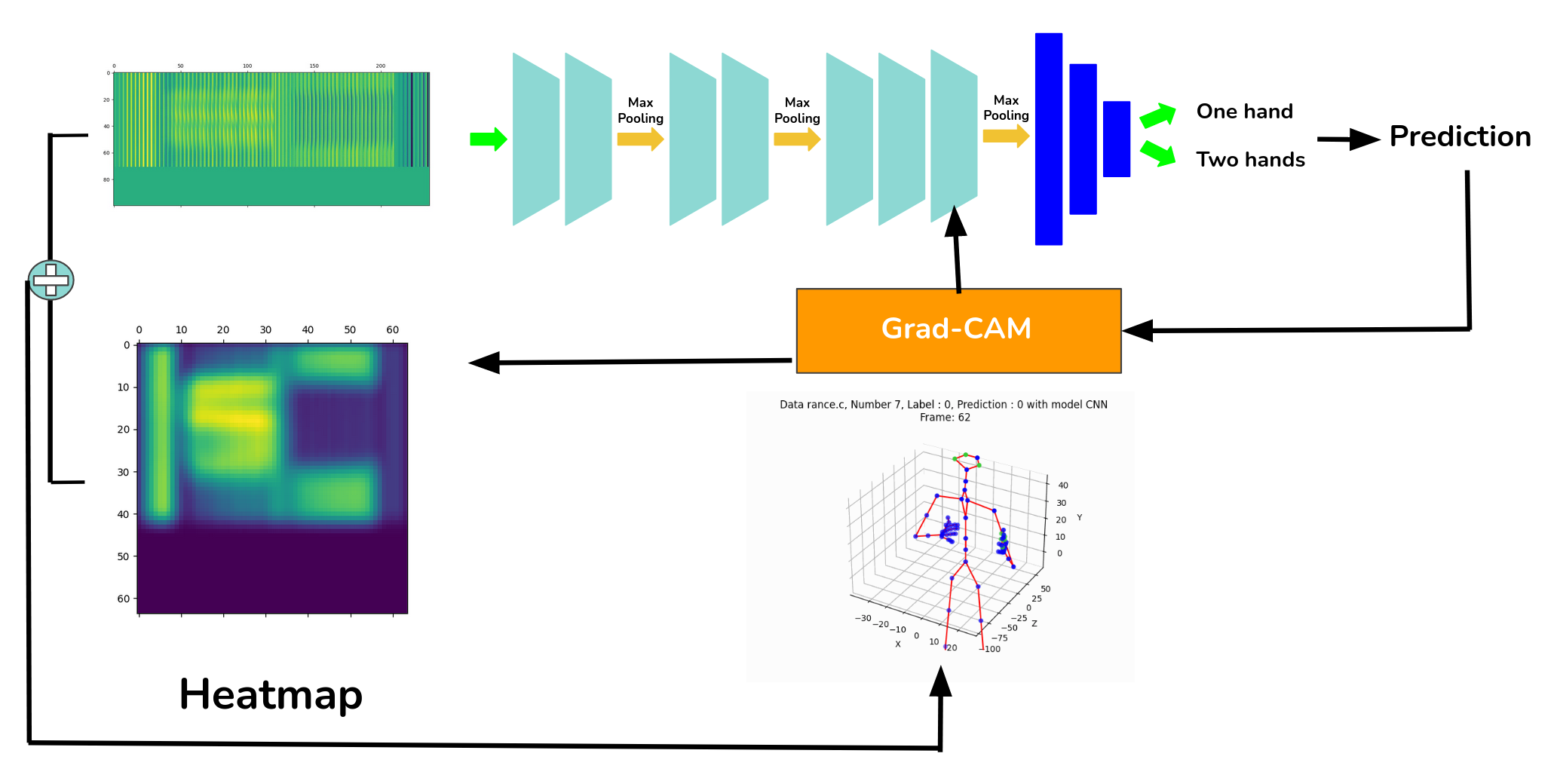}
    \caption{Grad-CAM applied to CNN for visualisation.}
    \label{gradcam}
\end{figure}

The main idea of the Grad-CAM method is to visualise which parts of the input data are more 'important' for predictions. The process starts with an input image, which in this case is the 2D array representing the values of the 3D coordinates of all 79 joints of skeleton for 100 frames. The image is passed through the layers of the CNN and the model makes a prediction. To understand why the model made a particular prediction, the following steps are involved :
\begin{itemize}
    \item Gradient computation: Computing the gradients of the predicted class score (in this case, one hand \textit{Mono} or two hands \textit{Bi}) with respect to the feature maps of the last convolutional layer.
    \item Global average pooling: Performing global average pooling on these gradients to obtain the weights for each feature map. These weights represent the importance of each feature map for the prediction.
    \item Weighted sum: Computing a weighted sum of the feature maps using the weights obtained in the previous step. This weighted sum highlights the important regions in the feature maps that contribute to the prediction.
    \item ReLU activation: Applying a ReLU activation function to the weighted sum to obtain the final Grad-CAM heat-map. The ReLU function ensures that only positive contributions are considered, highlighting the regions that positively influence the prediction.
\end{itemize}
The resulting heat-map is then overlaid on the input image (with interpolation to fit its size). To take the visualisation further, a max-pooling is done over the three values of the three dimensional coordinates of each joint on the heat-map, and from these values (one per joint), the ten most significant joints for each frame are extracted and highlighted in green in the animation of the skeleton. Figure \ref{gradcam} is an overview of the application of the Grad-CAM algorithm to the skeleton data.

\section{Task setting}

\subsection{Dataset}

\begin{figure}[H]
    \centering
    \includegraphics[width=15cm]{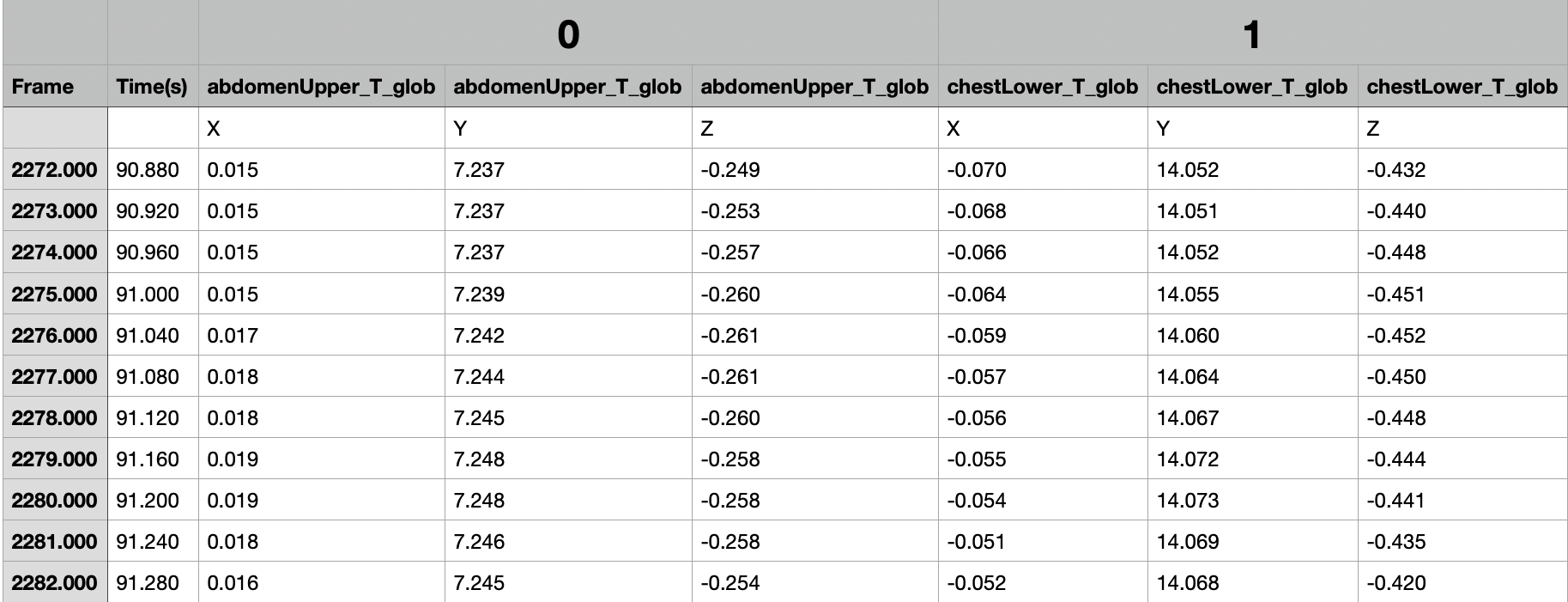}
    \caption{Example of skeleton file (\textsf{Avion.csv}).}
    \label{csvfile}
\end{figure}

The skeleton animation data used for this work was provided by MocapLab company \footnote{https://www.mocaplab.com/fr}, a high fidelity motion capture studio located in Paris. Each skeleton sequence consists of a \textsf{csv} file containing the three dimensional coordinates of each joint depicted in the skeleton over time. In other words, each sample can be seen as an array of \(3 \times n + 1\) columns and \(t\) rows where \(n\) is the number of joints in the skeleton and \(t\) is the duration of the movement. (The extra column contains the time stamp for each position). The first batch of files provided by MocapLab contained skeletons composed of \(n = 79\) articulations. Figure \ref{csvfile} shows the first ten time steps of the recording of a movement for the two first joints.

The dataset contained 111 skeleton files. An additional file containing the labels for each skeleton file was also provided by the motion capture studio. A skeleton file could be either labelled as \textit{Mono} or \textit{Bi} depending if the movement of the skeleton was a one handed or a two handed sign.

\subsection{Data Pre-processing}

\begin{figure}[]
    \centering
    \includegraphics[width=15cm]{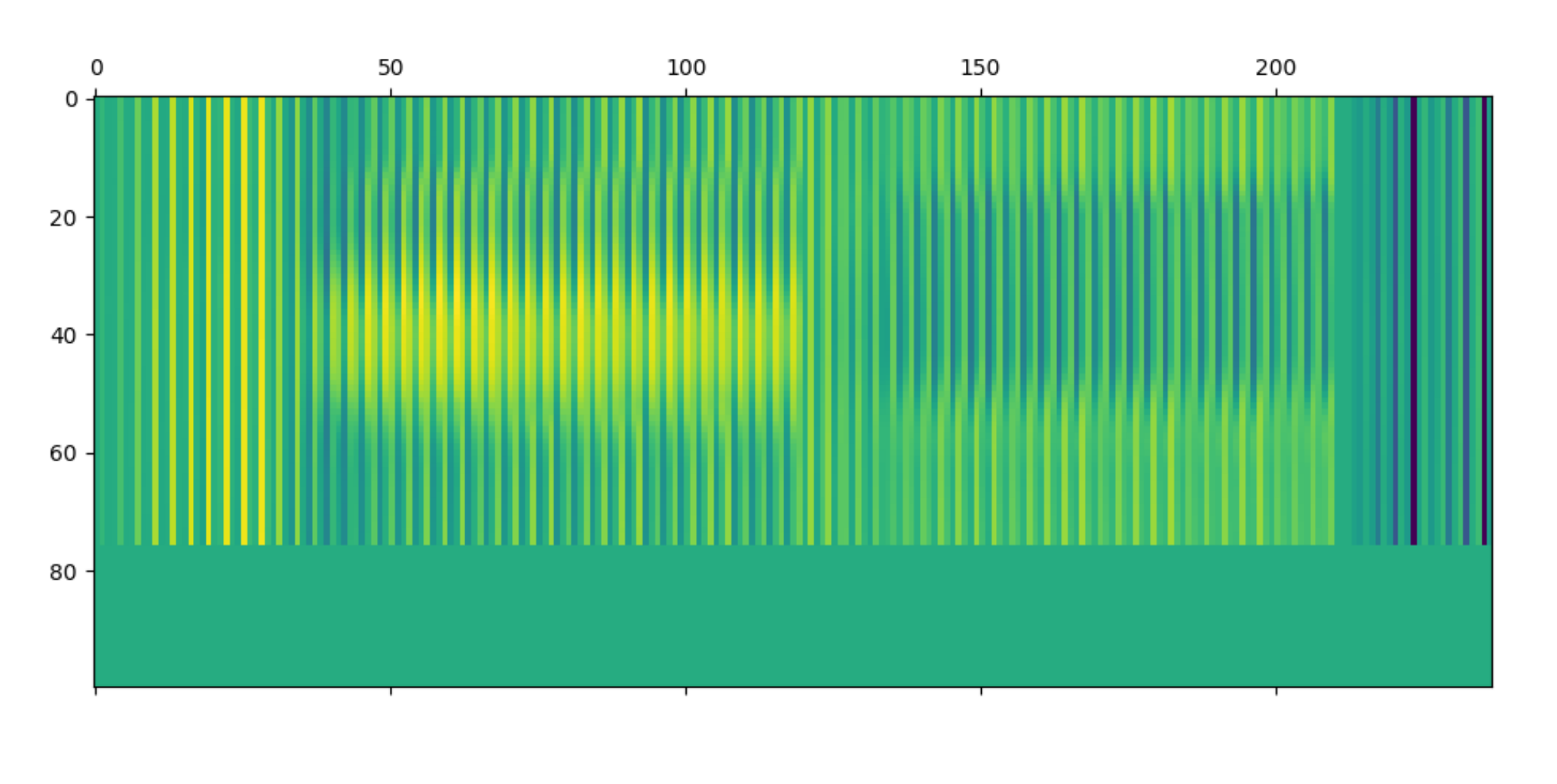}
    \caption{Examples of a matric of 3D skeleton data which has been transformed following our data pre-processing scheme.}
    \label{data_preprocessing}
\end{figure}

Before being fed to the different machine learning models, the skeleton files require some pre-processing. As all the models we tried required data of a fixed size, every file was extended (or padded) to match the length of the longest sequence. In other words, every skeleton file was transformed into an array of \(3 \times n + 1\) columns and \(t_{max}\) rows by adding some rows full of zeros, \(t_{max}\) being length of the longest skeleton file in the dataset. In the dataset we worked with, we found \(t_{max} = 100\). Figure \ref{data_preprocessing} shows the results of the pre-processing on a skeleton file. This two dimensional data sample could be fed to the CNN as is but the two other models (FC and LSTM) required the data to be flattened. Each skeleton file was then transformed into a vector of size \(3 \times n \times t_{max} = 237000\) (as shown in Figures \ref{model_FC}, \ref{model_LSTM}).

\subsection{Experiment Settings}

The target of the models is either 0 or 1, representing classes \textit{Mono} (one hand movement) and \textit{Bi} (two hands movement) respectively. The 111 labelled skeleton files available in the dataset were used as follows:
\begin{itemize}
    \item Supervised learning (all three models): 66 samples were used for training, 11 for validation and 34 testing.
    \item Self-supervised learning (applied to the FC and CNN models): 101 samples were used for the unsupervised training. In this simulated settings, the supervised model for the downstream task was trained with 5 samples for the train dataset and 5 samples for the validation dataset. The 101 data used for the unsupervised part were reused for the final test.
\end{itemize}
We used accuracy as the metric to evaluate each model. As a way to analyze and explain the results of the predictions, we applied the Grad-CAM method to our trained CNN in supervised learning. The source code of the methods (including the demonstration) can be found in this link \footnote{https://github.com/FABallemand/ProjetCassiopee}.

\section{Results}

\subsection{Supervised Learning}

According to our results, the three models we tested are able to learn properly to do the binary classification on the skeleton data. All of them can predict with high accuracy if a movement requires one or two hands. We observed a test accuracy of 100\% for the CNN and LSTM models. The FC model also performed very well with a test accuracy of 97\% as shown in Table \ref{supervised_results}. In this respect, it should also be noted that the time taken to train each model was under ten minutes, with the FC being the fastest model to train since it is the one containing the lowest number of parameters.

\begin{table}[]
    \centering
    \begin{tabular}{|l|c|c|c|}
        \hline
        \multicolumn{1}{|c|}{} & FC     & CNN    & LSTM   \\ \hline
        Test accuracy & 97\% & 100\% & 100\% \\ \hline
        Accuracy (Test+Validation) & 96.1\% & 97.4\% & 96.1\% \\ \hline
        Misclassified data & \textit{Venir}, \textit{Chambre}, \textit{Couche} & \textit{Venir}, \textit{Petit} & \textit{Venir}, \textit{Chambre}, \textit{Chef} \\ \hline
    \end{tabular}
    \caption{Results comparison for supervised learning on MocapLab skeleton data.}
    \label{supervised_results}
\end{table}

\begin{figure}[]
    \centering
    \subfloat[\centering Result of FC]{{\includegraphics[width=12cm]{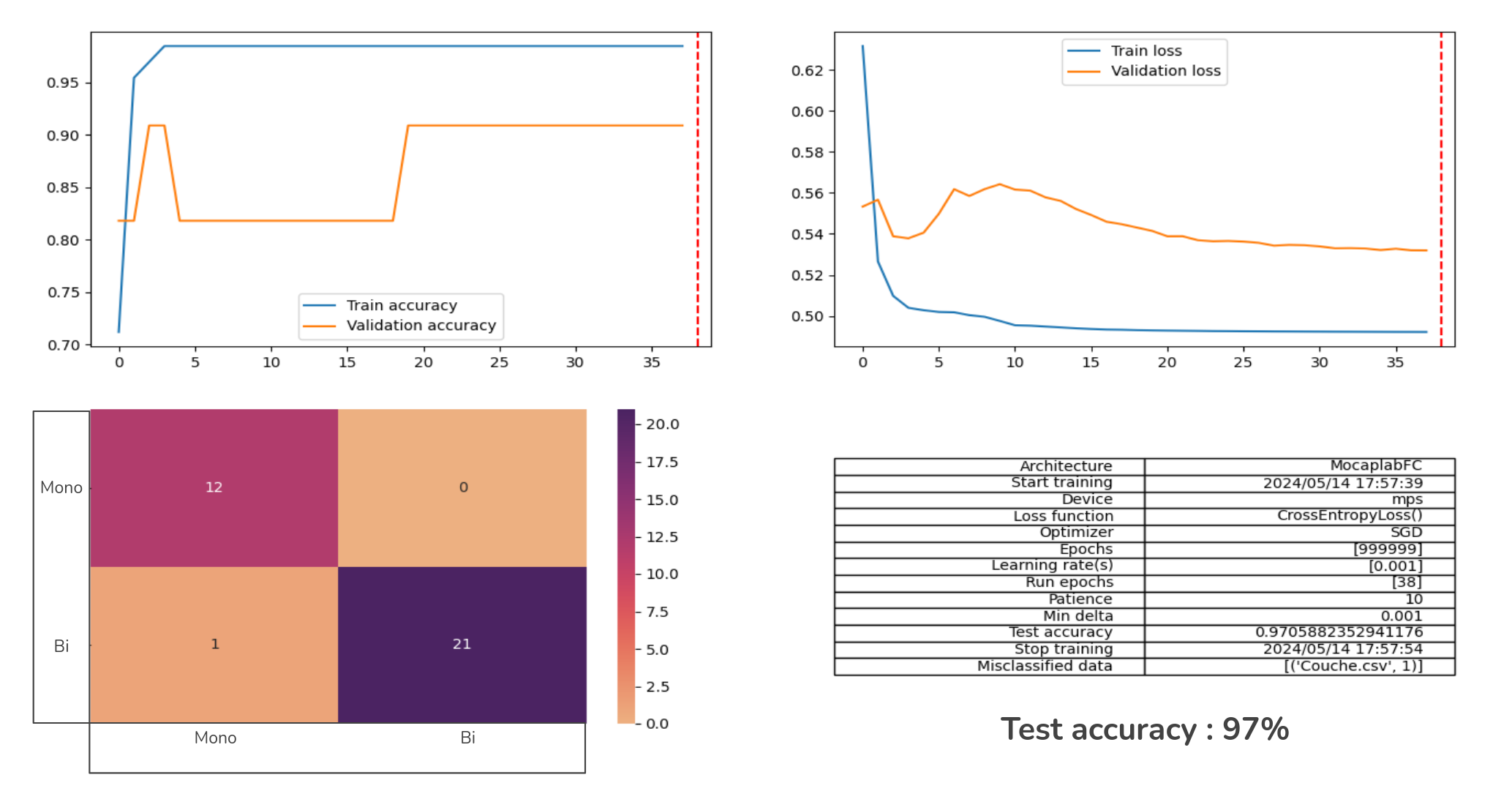} \label{results_FC}}}
    \qquad
    \subfloat[\centering Result of CNN]{{\includegraphics[width=12cm]{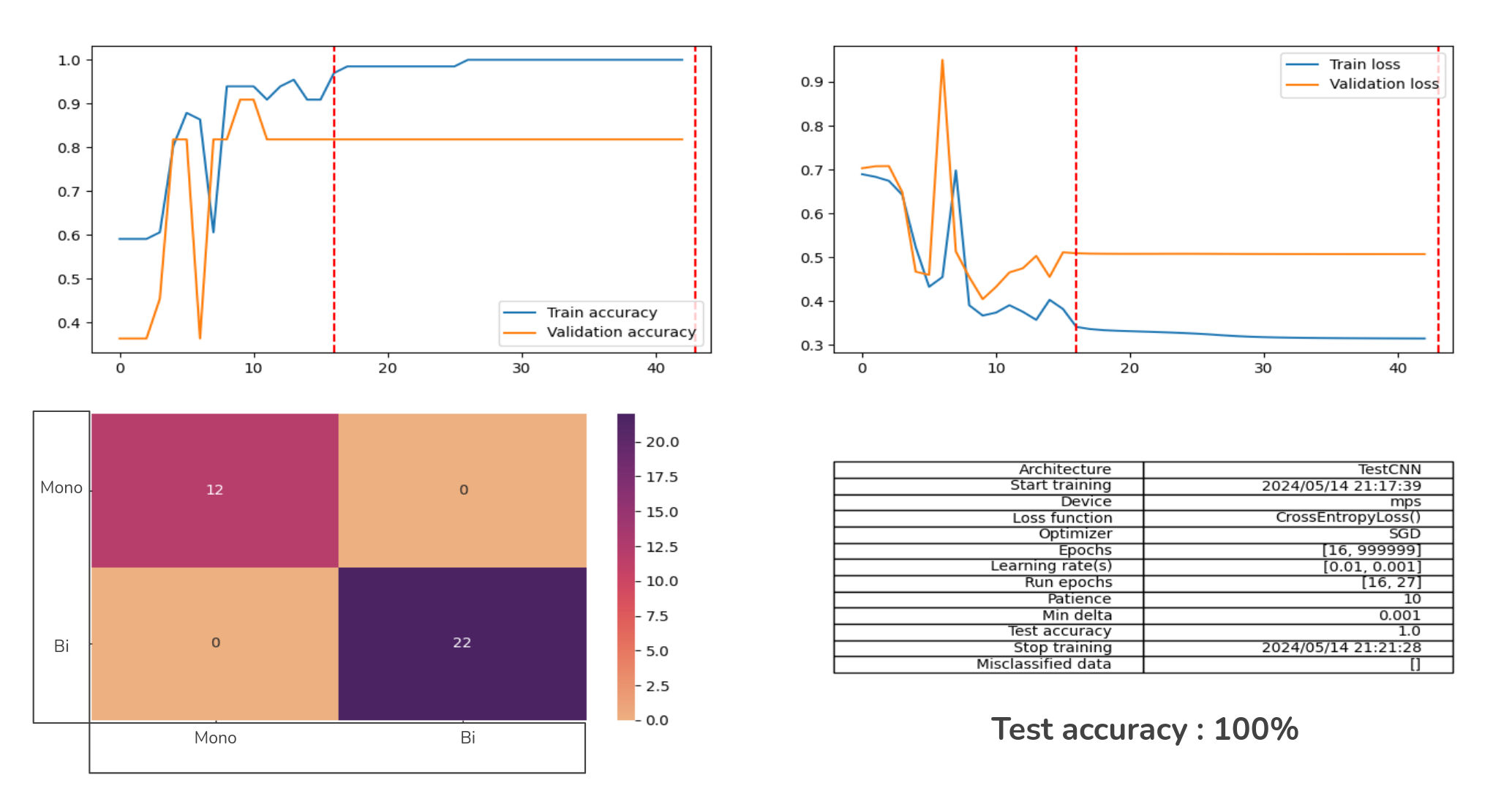} \label{results_CNN}}}
    \qquad
    \subfloat[\centering Result of LSTM]{{\includegraphics[width=12cm]{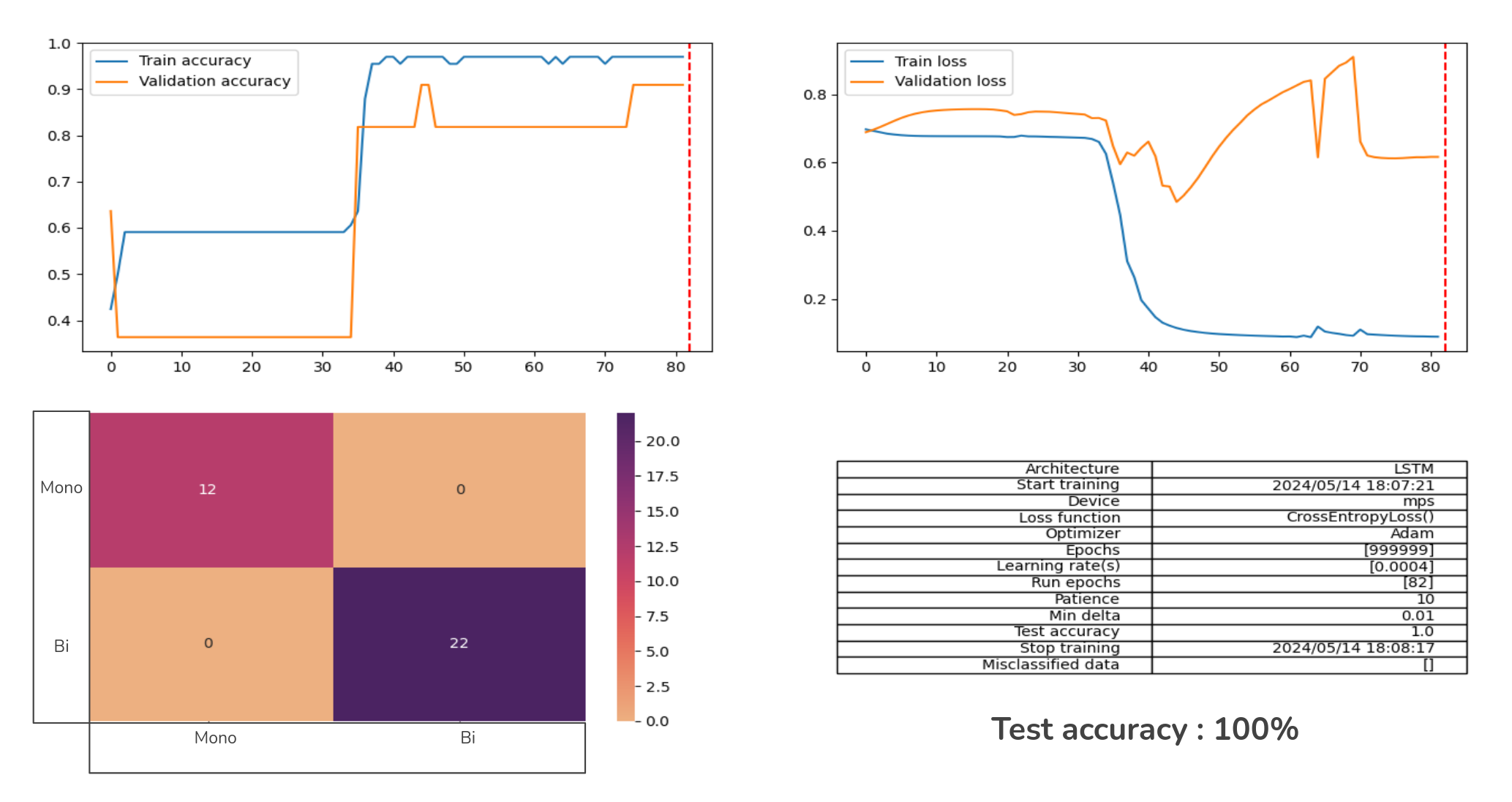} \label{results_LSTM}}}
    \caption{Training results (accuracy and loss graph for training and validation data during training, accuracy and confusion matrix on test data).}
    \label{results_graphs}
\end{figure}

The CNN has overall the best performances with only two misclassified data over the test and validation datasets, whereas the other models have three misclassified data (exhibited in Table \ref{supervised_results}). The loss curves in Figure \ref{results_CNN} suggest that the CNN is also the most stable model.

\begin{figure}[]
    \centering
    \subfloat[\centering Frame 22]
    {{\includegraphics[scale=0.24]{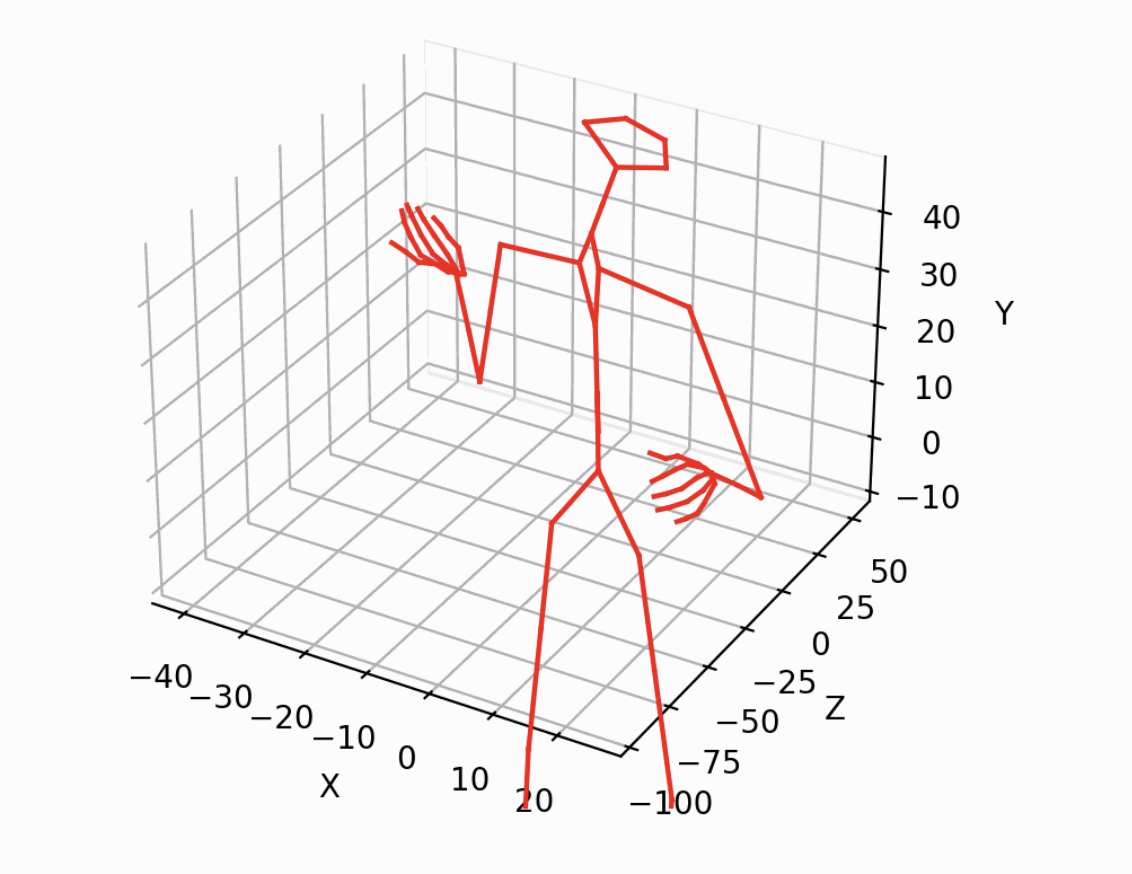} \label{venir_1}}}
    \qquad
    \subfloat[\centering Frame 28]
    {{\includegraphics[scale=0.24]{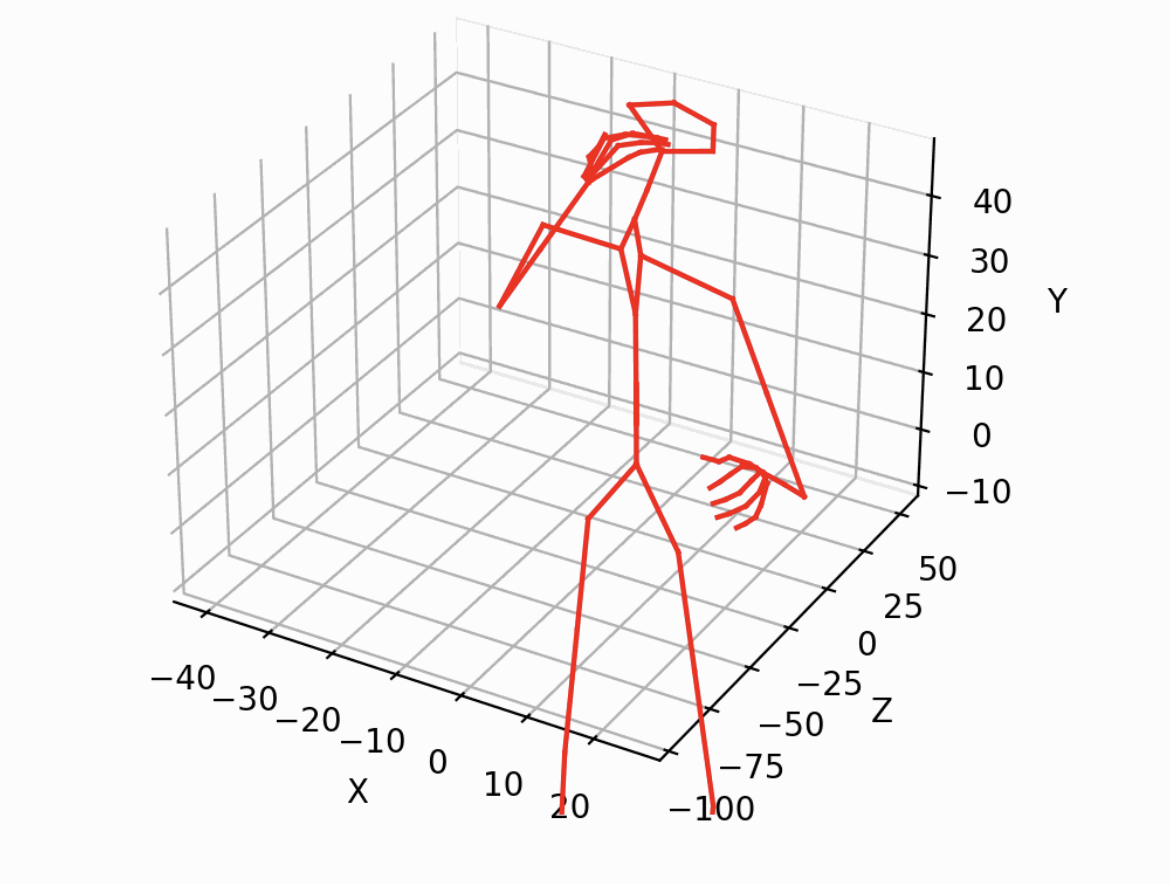} \label{venir_2}}}
    \qquad
    \subfloat[\centering Frame 35]
    {{\includegraphics[scale=0.24]{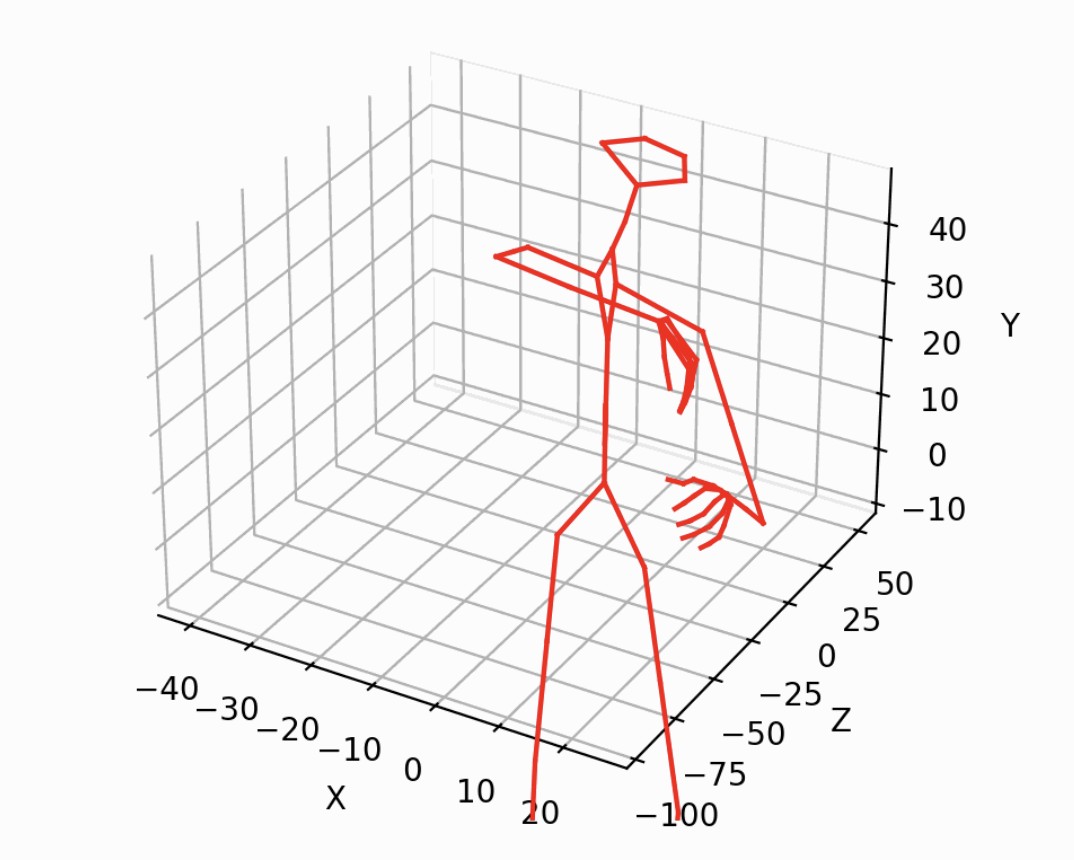} \label{venir_3}}}
    \caption{\textit{Venir} skeleton visualisation.}
    \label{venir}
\end{figure}

The data \textit{Venir} is misclassified by all three models (label 0, predicted as 1). One explanation could be that the movement uses only one hand but the unused hand is not as low as in the other movements. Additionally, the moving hand comes near the head for less than ten frames, meaning it is basically at the same height level as the unused hand for the most part of the gesture (cf. Figure \ref{venir}).

\subsection{Self-supervised Learning}
\label{SSL_experiement}

\begin{table}[]
    \centering
    \begin{tabular}{|c|c|c|}
        \hline
        Training method & Data utilisation & Test accuracy \\ \hline
        Supervised CNN (10\% of data) & 5\% train, 5\% validation, 90\% test & 83\% \\ \hline
        Self supervised CNN & 90\% for unsupervised learning, 5\% train, 5\% validation & 93\% \\ \hline
        {} & Note : data used for unsupervised training is also used for testing & {} \\ \hline
    \end{tabular}
    \caption{Result comparison for SL and SSL for the CNN model on Mocaplab skeleton data.}
    \label{ssl_mocaplab}
\end{table}

The experimentation with our three models has demonstrated the successful execution of binary gesture classification tasks. On this, we conducted an experiment using our CNN, training it with 5 data samples in both the training and validation datasets, while reserving the remaining 101 samples for testing. The results revealed a test accuracy of 83\% (Table \ref{ssl_mocaplab}), which, as anticipated, was lower than the previous outcomes.

Nonetheless, in the simulated scenario, self-supervised learning method can effectively capitalize on the scarcity of labelled data by using unlabelled data. We employed SSL to train our CNN, and achieved a test accuracy of 93\% (Table \ref{ssl_mocaplab}), which indeed increase the accuracy obtained as opposed to sole use of SL, hence this demonstrates the potential of SSL. Finally, other exploration experiment conducted on a larger dataset for multi-class object classification is presented in Appendix \ref{SSL_rgbd}.

\subsection{Visualisation of the models focus}

\begin{figure}[]
    \centering
    \subfloat[\centering visualisation for \textit{Mono}.]{{\includegraphics[width=15cm]{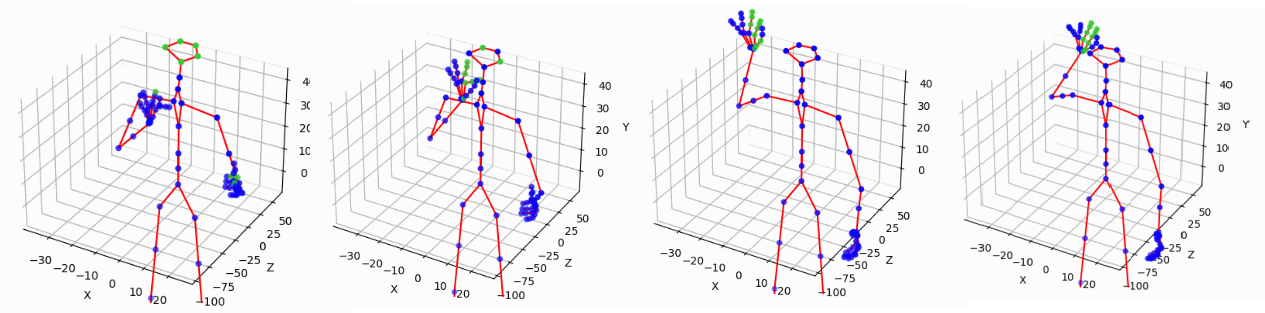} \label{gradcam_mono}}}
    \qquad
    \subfloat[\centering visualisation for \textit{Bi}.]{{\includegraphics[width=15cm]{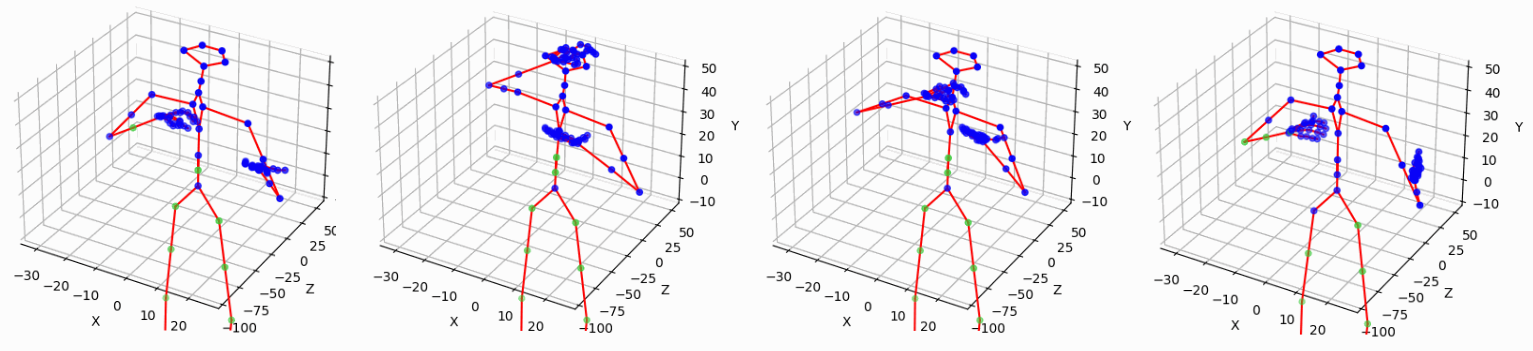} \label{gradcam_bi}}}
    \caption{Grad-CAM based visualisation of models focus with respect to the joint coordinates (green dots are the point where models focus) for one and two hands sign.}
    \label{gradcam_mono_bi}
\end{figure}

In Figure \ref{gradcam_mono_bi}, we can see two examples of Grad-CAM visualisation, \ref{gradcam_mono} is for a one hand sign (\textit{Mono} label) and \ref{gradcam_bi} is for a two hands sign (\textit{Bi} label). The example for \textit{Mono} shows that the CNN focuses on the joints of the moving hand to make its prediction. Most of the one hand sign data show similar patterns. The fact that the model performs the prediction by analyzing the joints in the moving hands (joints) provides some insights where the model is focusing (i.e. the relevant joints) \cite{zhang2022predicting}. However, all two hands signs Grad-CAM visualisation results are identical as shown on the Figure \ref{gradcam_bi}, meaning the focus of the model is focusing on the static area (such as legs and torsos). As such, improving the explainability of the predictions of the CNN for \textit{Bi} labelled data can be considered as a relevant task in the future.

\section{Conclusion}

In this work, we explore the application of deep self-supervised learning to the field of automatic gesture recognition, specifically focusing on hand gestures using 3D moving skeleton data. Our objective is to assess the performance of various neural network architectures — fully connected network, convolutional neural networks, and long short-term memory networks — to recognize gestures by analyzing 3D joint coordinates over time. This approach provides a comprehensive representation of motion, which is crucial for accurately capturing and understanding complex gestures.

Our findings indicate that supervised learning techniques yield high accuracy in gesture classification with this data type, where CNN and LSTM models achieve perfect test accuracy of 100\%, and the FC model follows closely with 97\% accuracy. The CNN model demonstrate superior performance, particularly in handling the temporal and spatial complexities of 3D skeleton data, making it well-suited for the task.

The application of self-supervised learning methods shows promising potential, particularly in scenarios with limited labelled data. By pre-training models on simulated task of using unlabelled 3D skeleton data, we observe enhanced performance when transitioning to supervised tasks. Although the results for self-supervised learning are slightly less conclusive, they underscore the importance of leveraging large amounts of unlabelled data to improve model training efficiency and accuracy.

Our experiments with Grad-CAM provide valuable insights into model interpretability and understanding the decision-making process of our CNN model, particularly for gestures involving single-hand movements. However, the visualisation for two-handed gestures indicates a need for further refinement in model interpretability, as the focus tends to misaligned with the intended regions of interest.

Moving forward, future work could explore more complex classification tasks involving multiple gesture categories and a larger dataset to improve generalization and model robustness. Additionally, investigating advanced self-supervised learning techniques and integrating multi-modal data sources could further enhance the performance of gesture recognition systems, broadening their application in areas such as sign language recognition, virtual reality, healthcare, and human-computer interaction.

As such, this work demonstrates the significant potential of deep learning and self-supervised learning techniques in advancing automatic gesture recognition, paving the way for more accessible and intuitive human interactions and human-computer interfaces.

\section*{Acknowledgments}

 We would like to thank Zied LAHIANI at Department of ARTEMIS for the technical support. The authors would also like to thank Remi BRUN and Boris DAURIAC from Mocaplab studio for providing the captured high quality data which is used for study case on this work.

%Bibliography
\bibliographystyle{unsrt}  
\bibliography{references}  
\newpage

%\appendix
\begin{appendices}
\section{Self-supervised Learning for Object Classification}
\label{SSL_rgbd}

\begin{figure}[h!]
    \centering
    \includegraphics[width=12cm]{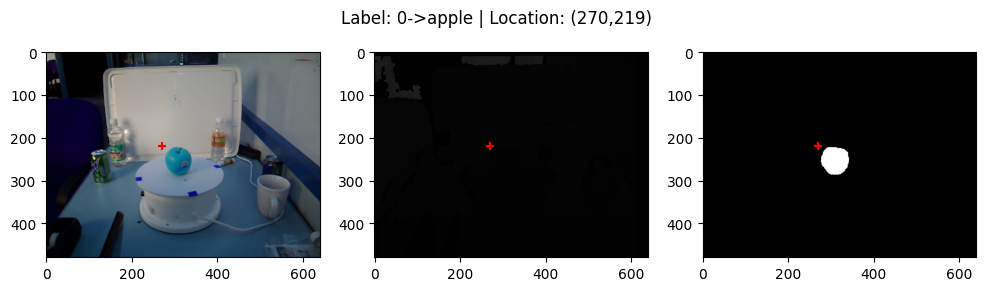}
    \caption{Data sample from the RGB-D Object dataset after pre-processing step.}
    \label{rgbd_object_sample}
\end{figure}

The experiment from Section \ref{SSL_experiement} shows that unlabelled data can be leveraged in the context of hand gesture recognition thanks to SSL. Having at our disposal only a few skeleton files labelled for binary classification, we also wanted to know if SSL could be scaled up with more data and for multi-class classification. For that matter, we used the \href{https://rgbd-dataset.cs.washington.edu/dataset/rgbd-dataset_full/}{RGB-D Object Dataset}, a very large dataset of RGB, depth and segmented images (each of a resolution of 640x480 as shown in Figure \ref{rgbd_object_sample}) with object class and position labels. This dataset contains more than 124000 data samples distributed across 53 object classes. We implemented SL for object classification, UL for image reconstruction (with and without contrastive learning and SSL with object classification as the downstream task, as described in Section \ref{SSL_section}, using only RGB images and class labels. We used an auto-encoder \cite{ae_1, ae_2} for image reconstruction during the unsupervised task with ResNet-18 \cite{he2016deep} as the encoder part. The same pre-trained ResNet-18 was then fine tuned and used for the downstream task of image classification.

The results presented in Table \ref{results_RGBD} show that SL achieve great to moderate accuracy results when trained on the entire or on half of the dataset. However, the results of SSL are below our expectations. While there is no noticeable accuracy loss, results of SSL are on par with the results obtained with SL with an accuracy of around 50\%. In this particular case, the effect of SSL still is not achieved as larger training time and more (unlabelled) data than SL are required. This thus yields the slightly reduced accuracy obtained.

It should be noted that this behavior may be due to our implementation of SSL on this dataset. While our model has a lot in common with the SimCLR framework presented in \cite{chen2020simple}, our implementation is different notably at the level of the contrastive learning. Our implementation relies on labels in order to select data samples similar or different than the sample currently processed whereas SimCLR consider every data sample of a batch to be different from the others. Our implementation should enhance the separability of workion in the latent space but it requires labels which makes it unsuitable if the goal is to leverage large amount of accessible unlabelled data. Due to some technical difficulties and a lack of time, we did not investigate the issue further. However, we believe that this task can be further be explored to reveal more result.

\begin{table}[h!]
\begin{tabular}{|l|l|c|c|cc|}
\hline
\multicolumn{1}{|c|}{\multirow{2}{*}{}} & \multicolumn{1}{c|}{\multirow{2}{*}{Training Method}} & \multirow{2}{*}{Model}                                                         & \multirow{2}{*}{Data} & \multicolumn{2}{c|}{Metric}                 \\ \cline{5-6} 
\multicolumn{1}{|c|}{}                  & \multicolumn{1}{c|}{}                                 &                                                                                &                       & \multicolumn{1}{c|}{Accuracy (\%)} & F1     \\ \hline
1                                       & Supervised Learning                                   & Pretrained ResNet18                                                            & Split A + B           & \multicolumn{1}{c|}{93.47}         & 0.9320 \\ \hline
2                                       & Supervised Learning                                   & Pretrained ResNet18                                                            & Split A               & \multicolumn{1}{c|}{49.15}         & 0.3540 \\ \hline
3                                       & Unsupervised Learning                                 & \begin{tabular}[c]{@{}c@{}}Autoencoder with\\ pretrained ResNet18\end{tabular} & Split B               & \multicolumn{1}{c|}{}              &        \\ \hline
4                                       & Unsupervised Learning with Contrastive Loss           & \begin{tabular}[c]{@{}c@{}}Autoencoder with\\ pretrained ResNet18\end{tabular} & Split B               & \multicolumn{1}{c|}{}              &        \\ \hline
5                                       & Self-supervised Learning                              & \begin{tabular}[c]{@{}c@{}}Pretrained ResNet18\\ from 3\end{tabular}           & Split A               & \multicolumn{1}{c|}{48.29}         & 0.3412 \\ \hline
6                                       & Self-supervised Learning                              & \begin{tabular}[c]{@{}c@{}}Pretrained ResNet18\\ from 4\end{tabular}           & Split A               & \multicolumn{1}{c|}{48.37}         & 0.3446  \\ \hline
\end{tabular}
\caption{Results comparison for SL and SSL for object classification on RGB data from the RGB-D Object dataset.}
\label{results_RGBD}
\end{table}

\end{appendices}

\end{document}